\documentclass{article}
\usepackage{PRIMEarxiv}

\usepackage[utf8]{inputenc} 
\usepackage[T1]{fontenc}    
\usepackage{hyperref}       
\usepackage{url}            
\usepackage{booktabs}       
\usepackage{amsfonts}       
\usepackage{nicefrac}       
\usepackage{microtype}      
\usepackage{lipsum}
\usepackage{fancyhdr}       
\usepackage{graphicx}       
\graphicspath{{media/}}     
\usepackage{listings}
\title{Multimodal Datasets and Benchmarks for Reasoning about Dynamic Spatio-Temporality in Everyday Environments}

\usepackage{graphicx}
\usepackage{listings}
\usepackage{hyperref}

\author{Takanori Ugai\\
{\small Fujitsu Limited.}\\
{\small 4-1-1 Kamikotanaka Nakaharaku Kawasaki Kanagawa, 211-8588, Japan}\\
{\tt\small ugai@fujitsu.com}\\
{\small National Institute of Advanced Industrial Science and Technology}\\
{\small 2-3-26, Aomi, Koto-ku, Tokyo 135-0064, Japan}
\and
Kensho Hara, Shusaku Egami, Ken Fukuda\\
{\small National Institute of Advanced Industrial Science and Technology}\\
{\tt\small \{kensho.hara, s-egami, ken.fukuda\}@aist.go.jp}
}

\begin{document}
\maketitle
\begin{abstract}

We used a 3D simulator to create artificial video data with standardized annotations, aiming to aid in the development of Embodied AI. Our question answering (QA) dataset measures the extent to which a robot can understand human behavior and the environment in a home setting. Preliminary experiments suggest our dataset is useful in measuring AI's comprehension of daily life.
\end{abstract}

\section{Introduction}
\label{sec:intro}
\vspace{-0.3em}
As Embodied AI continues to develop, understanding the time and place of actions in daily life becomes increasingly important ~\cite{RePEc:nat:nature:v:588:y:2020:i:7837:d:10.1038_d41586-020-03412-z, saycan2022arxiv, pmlr-v164-blukis22a, paolo2024embodied,5514606}.
Datasets and benchmarks have been created to support their development, and challenges have been presented \cite{deitke2022retrospectives,9879279, ning2023videobench}.

Most of this data consists of recorded images of everyday life, annotations, and descriptions.
The annotations were performed manually and were imprecise; not everything in the room was annotated.
The behavior of what the person tries to do needs to be fully described in these descriptions.


Nishimura et al.~\cite{published_papers/36290329} proposed PrimitiveActionOntology\footnote{https://github.com/aistairc/PrimitiveActionOntology} to abstract activity labels in recognition datasets based on HomeOntology~\cite{10.1007/978-3-030-59833-4_3} and International Classification of Functioning, Disability and Health (ICF)\footnote{https://www.who.int/standards/classifications/international-classification-of-functioning-disability-and-health}. They also proposed a HomeObjectOntology\footnote{https://github.com/aistairc/HomeObjectOntology} based on VirtualHome assets, objects defined in Charades~\cite{10.1007/978-3-319-46448-0_31}, and objects that occurred in the videos in the video archive called Elderly Behavior Library\footnote{https://www.behavior-library-meti.com/behaviorLib/homes/about}.

We created artificial video data ({\bf MMDL}: Multimodal Dataset of Daily Life) using a 3D VirtualHome-AIST~\cite{ugai2024synthetic} simulator, which is based on VirtualHome~\cite{virtualhome} and, using VirtualHome2KG~\cite{published_papers/41642998}, created data describing what it is and where it is located for more objects.
These data also clarify what the data are from the scripts that are placed in the simulator to make the avatar work.
The annotations are mechanically generated with a standard vocabulary based on PrimitiveActionOntology and HomeOntology, which contributes significantly to the development of Embodied AI as they are consistent and free of contradictions.

We also created a question answering (QA) dataset ({\bf MMQADL}: Multimodal Question Answering Dataset of Daily Life) to measure the extent to which the robot could understand a person's daily life from a video.
We offer various types of descriptive and quantitative questions for question answering (QA) to gather information on location, action, object, time, and more. We also provide location-selective and descriptive QA examples for training and evaluation data.

This paper presents the findings of initial experiments conducted using two generative AIs, namely Video-LLaVa~\cite{lin2023video} and Google's Gemini 1.5 Pro Vision. These AIs were fed with a combination of images, natural language sentences, and QA that we created. The purpose of this experiment was to investigate AI's understanding of human behavior in a home environment. The results of the experiment indicate that our dataset is useful in measuring the AI's comprehension of human behavior and the surrounding environment in a home.

\section{MMDL: Simulation movie and detailed annotation}

\begin{figure}[th]
 \centering
 \includegraphics[width=0.9\linewidth]{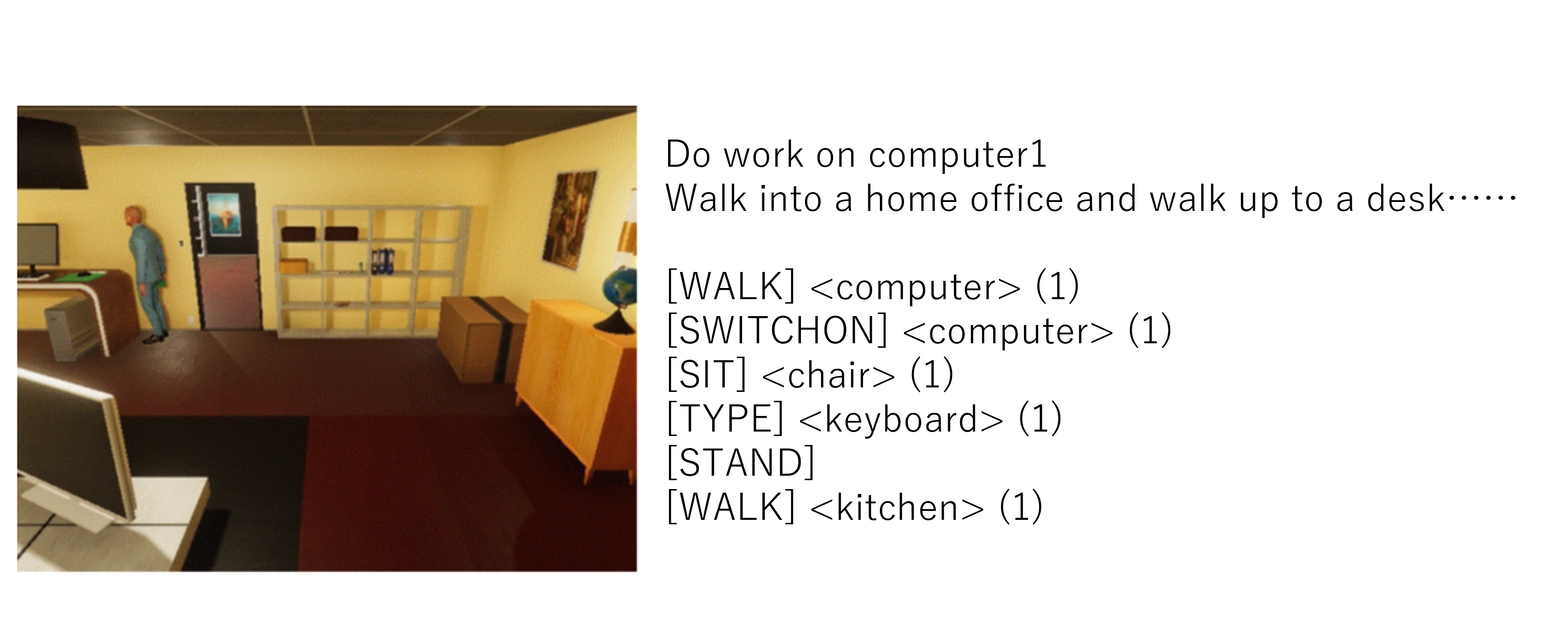}
 \caption{Example of video snapshot and action script}
 \label{fig:event_class_diagram}
\end{figure}

Figure \ref{fig:event_class_diagram} shows an action script titled ``Do work on computer'' and a snapshot of the video generated from it in VirtualHome-AIST. The first line of the action script is the title, the second line is the description, and the fourth and subsequent lines are the rows of the avatar's behavior.
There are 3,530 different videos, each of which shows a short chunk of behavior, called an activity, of approximately 30 seconds to a minute in length. They were generated from 706 scenarios (action scripts); for one scenario, five videos were generated with different camera positions.
The characters' behavior (avatars) in the videos and the 3D coordinates of approximately 400 objects in the house were annotated as data using VirtualHome2KG. The states of lights and other electrical appliances, such as on/off, and the opening/closing of the fridge and room doors, were also recorded~\cite{ugai2024synthetic}, and the 2D positions of the camera image were also annotated for the objects with which the avatar was involved. 2D annotation is provided in the same scene graph format as Action Genome~\cite{Ji_CVPR_2020}.

\section{MMQADL: QA dataset for measuring daily life understanding}

\begin{lstlisting}[caption = Example of Question and Answer , label = program1]
Q: Where is the man 10 seconds later from the 
beginning of the video?
A1: Livingroom
A2: Bedroom
A3: Kitchen
A4: Bathroom
\end{lstlisting}

QA can pose different types of questions. They can be a choice (Listing 1) or a simple ``yes'' or ``no'' answer.
The questions were designed to gather information about the location, action, object, time, and combination of topics being discussed based on TempCompass~\cite{liu2024tempcompass} and MVBench~\cite{Li2023MVBenchAC}. In addition, there are questions that focus on the appropriate caption for a video, which can be either short or long.

These questions were classified into two types: descriptive and quantitative. Descriptive questions were used to obtain factual information or details about the topic, event, object, or situation. They typically start with words like ``what,'' ``does,'' ``where,'' and ``when.'' The quantitative questions were designed to obtain numerical or quantitative data. They typically start with words like ``how many,'' ``how much,'' ``how long,'' and ``how often.''

The data were designed to provide answers to 70 types of questions for longer time frames, defined in columns of three to seven activities, and provided in JSON format. The QA data were divided into two parts: learning (80\%) and evaluation (20\%) data, both containing answers.

The training data not only lacks answers but also provides annotated data with missing locations, actions, and objects that correspond to the answers. Additionally, the questions were categorized as Easy or Hard. Easy questions have only two options, while hard questions have around 30 options for actions, and all objects (about 200) that exist in the house are considered candidates for objects.

\section{Preliminary Experiment}

\begin{table}
\caption{Score of Precision}\label{result}
\centering
\begin{tabular}{l|ccccc}  
  & \textbf{action} & \textbf{location} & \textbf{object} & \textbf{time} & \textbf{caption} \\ \hline
  {Gemini} & 0.7 & 0.9 & 0.4 & 0.5 & 0.8 \\
  {Video-LLaVa} & 0.5 & 0.4 & 0.25 & 0.1 & 0.6
\end{tabular}
\end{table}

In the Knowledge Graph Reasoning Challenge 2024, one of the strategies~\cite{10475623} is to complete the annotations provided in the Knowledge Graph from the video. If we can accurately fill in all the missing parts of the annotation, we can answer all the questions correctly. We have used video clips and questions about the missing actions, locations, objects, and time as input. We tested the Large Language Models to answer the questions, and Table \ref{result} is the result of our experiment. Overall, Gemini performs well. In particular, the distinction between the four types of rooms is mostly accurate. Video-LLaVa, on the other hand, does not understand the time elapsed in the video.

\section{Summary}

This article discusses the creation of a dataset that supports the development of Embodied AI. The dataset includes artificial movie data and a QA dataset to measure the AI's comprehension of human behavior in a home environment. The results of the initial experiments show that the dataset is useful for measuring the AI's understanding of human behavior and the surrounding environment in a home.
We are planning to organise a technology contest (Challenge) in the future.
All data is publicly available from \url{https://github.com/KGRC4SI/DataSet}

\section*{Acknowledgement}
This paper is based on results obtained from projects, JPNP20006 and JPNP180013, commissioned by the New Energy and Industrial Technology Development Organization (NEDO).

\bibliographystyle{ACM-Reference-Format}
\bibliography{main}
\end{document}